# A Latent Variable Approach to Gaussian Process Modeling with Qualitative and Quantitative Factors

Yichi Zhang[1], Siyu Tao[1], Wei Chen[1], and Daniel W. Apley[2]

[1]Department of Mechanical Engineering, Northwestern University, Evanston, IL

[2]Department of Industrial Engineering and Management Sciences, Northwestern University, Evanston, IL

## Abstract

Computer simulations often involve both qualitative and numerical inputs. Existing Gaussian process (GP) methods for handling this mainly assume a different response surface for each combination of levels of the qualitative factors and relate them via a multiresponse cross-covariance matrix. We introduce a substantially different approach that maps each qualitative factor to an underlying numerical latent variable (LV), with the mapped value for each level estimated similarly to the correlation parameters. This provides a parsimonious GP parameterization that treats qualitative factors the same as numerical variables and views them as effecting the response via similar physical mechanisms. This has strong physical justification, as the effects of a qualitative factor in any physics-based simulation model must *always* be due to some underlying numerical variables. Even when the underlying variables are many, sufficient dimension reduction arguments imply that their effects can be represented by a low-dimensional LV. This conjecture is supported by the superior predictive performance observed across a variety of examples. Moreover, the mapped LVs provide substantial insight into the nature and effects of the qualitative factors.

# 1. INTRODUCTION

Computer simulations play essential roles in today's science and engineering research. As an alternative to more difficult and expensive physical experiments, computer simulations help to explore or experiment with the physical process and understand how input factors affect a response of interest. Gaussian process (GP) models, a.k.a., kriging models have become the most popular method for modeling simulation response surfaces (Fang et al. 2006; Sacks et al. 1989; Santner et al. 2003). These standard methods for the design and analysis of computer experiments were developed under the premise that all the input variables are quantitative, which fails to hold in many applications. For example, consider a stamping operation, in which the response is the maximum strain over a stamped panel, and one of the factors affecting strain is the qualitative factor lubricant type (e.g., three different types: A, B, and C). Another example is the fluid dynamics model for investigating the thermal dynamics of a data center (Qian et al. 2008), which involves qualitative factors such as "hot air return vent location" and "power unit type".

To make the discussion more concrete, let $y(\cdot)$ denote the computer simulation response model with inputs $\boldsymbol{w} = (\boldsymbol{x}, \boldsymbol{t}) \in \mathbb{R}^{p+q}$, where $\boldsymbol{x} = (x_1, x_2, \ldots, x_p)$ represents $p$ quantitative variables, and $\boldsymbol{t} = (t_1, t_2, \ldots, t_q)$ represents $q$ qualitative factors, with the $j$th qualitative factor having $m_j$ levels, $j = 1,2,\ldots,q$. For simplicity, when constructing the relationship between $y$ and quantitative inputs $\boldsymbol{x}$, consider the GP model

$$y(\boldsymbol{x}) = \mu + G(\boldsymbol{x}), \tag{1}$$

where $\mu$ is the constant prior mean, $G(\boldsymbol{x})$ is a zero-mean GP with covariance function $K(\cdot,\cdot) = \sigma^2 R(\cdot,\cdot)$, $\sigma^2$ is the prior variance, and $R(\cdot,\cdot\,|\boldsymbol{\phi})$ denotes the correlation function with parameters $\boldsymbol{\phi}$. A commonly used correlation function for quantitative variables is the Gaussian correlation function

$$R(\boldsymbol{x}, \boldsymbol{x}') = \exp\left\{-\sum_{i=1}^{p} \phi_i (x_i - x_i')^2\right\}, \qquad (2)$$

which represents the correlation between $G(\boldsymbol{x})$ and $G(\boldsymbol{x}')$ for any two input locations $\boldsymbol{x} = (x_1, \dots, x_p)$ and $\boldsymbol{x}' = (x_1', \dots, x_p')$, where $\boldsymbol{\phi} = (\phi_1, \dots, \phi_p)^T$ is the vector of correlation parameters to be estimated via MLE, along with $\mu$ and $\sigma^2$. The correlation between $y(\boldsymbol{x})$ and $y(\boldsymbol{x}')$ depend on the spatial distance between $\boldsymbol{x}$ and $\boldsymbol{x}'$ and the correlation parameters. Other choices of correlation functions include power exponential, Matèrn (Rasmussen et al. 2006), and lifted Brownian (Plumlee and Apley 2017).

These types of correlation functions cannot be directly used with qualitative factors because the distances between the levels of qualitative factors is not defined. To incorporate both qualitative and quantitative factors into GP modeling, one must construct some appropriate correlation structure that is applicable over the qualitative factors. A method using a quite general and unrestrictive correlation structure for qualitative factors was developed in Qian et al. (2008), which treats the computer model as a multi-response GP with a different response for each combination of levels of the qualitative factors and then models the cross-correlation between different levels of the qualitative factors. As originally proposed, their method requires a rather complex optimization procedure to ensure that the correlation matrix is positive definite. Qian et al. (2008) also considered a number of special cases to simplify the structure. In Zhou et al. (2011), the authors employed hypersphere decomposition (Rebonato and Jäckel 1999) to model the correlation of qualitative factors, which significantly simplified the estimation procedure. Zhang and Notz (2015) showed that a certain representation of the qualitative factors via indicator variables results in the Qian et al. (2008) covariance model and considered a number of special cases. An advantage of the indicator variable approach is that it allows one to handle qualitative factors using standard GP modeling software developed for quantitative inputs. Correlation models with restrictions on the cross-correlation structure between the different factor levels were also proposed in (Joseph and Delaney 2007; McMillan et al. 1999; Qian et al. 2008). These can significantly reduce the complexity of the model, while sacrificing some flexibility for capturing various types of correlation structures for the qualitative factors. Recently, for multiple qualitative factors, Deng et al. (2017) proposed an additive GP model that is the sum of $q$ independent GPs over $(x, t_j)$ for $j = 1,2,\ldots,q$, assuming each GP has separable correlation in $x$ and $t_j$. We discuss the above methods in more detail in Section 3.

In this paper, we propose a fundamentally different method of handling qualitative factors in GP models that involves a latent variable representation of the qualitative factors. The fundamental idea behind the method is to map the levels of each qualitative factor to a set of numerical values for some underlying latent unobservable quantitative variable(s). After obtaining this mapping, our GP covariance model over $(x, t)$ can be any standard GP covariance model for quantitative variables over $(x, z(t))$, where $z(t)$ is the numerical vector of mapped latent variables. It is important to note that the mapped values $\{z(t)\}$ are obtained in a straightforward and computationally stable manner via maximum likelihood estimation (MLE) along with the correlation parameters for $x$, and the mapping is scaled so that the correlation parameters for $z$ are unity. In Section 4, we show that over a broad array of mathematical and engineering examples, a two-dimensional latent variable space for each qualitative factor is sufficient and flexible enough to represent many correlation structures. As a beneficial side-effect, the latent variable mapping provides an inherent ordering and structure for the levels of the qualitative factor(s), which can provide substantial insight into the effects of the qualitative factors. We also demonstrate this in the examples.

There are strong physical arguments for why our mapped latent variable approach constitutes a covariance parameterization that, while tractable and involving relatively few parameters to estimate, is flexible enough to capture the behavior of many real physical systems. For any real physical system model having a qualitative input factor, there are *always* underlying physical variables that account for differences in the response across the different levels of the factor. For example, in the earlier stamping example, differences in the response (panel deformation and strain behavior) due to the different lubricant types *must* be due to the lubricant types having different underlying physical properties, such as lubricity, viscosity, density, thermal stability of the oil, etc. Otherwise, there is no way to code a simulation model to account for the effects of lubricant type. Thus, the lubricant type can indeed be represented by some underlying numerical variable(s).

There may be many underlying variables, but the Pareto principle implies that their collective effect may often be attributed to just a few variables. Even without invoking the Pareto principle, if there are many underlying latent variables associated with the factor levels that effect the response, their collective effect on the response will usually be captured by some low-dimensional latent combination of the variables. To see this, suppose the effect of lubricant type $t$ on the response was due to 20 underlying numerical variables $\{v_1(t), v_2(t), \ldots, v_{20}(t)\}$ associated with the type, in which case their collective effect on the response can be written as $y = g(\boldsymbol{x}, v_1, \ldots, v_{20})$ for some function $g(\cdot)$. If, for example, the dependence happens to be of the form $y \cong g(\boldsymbol{x}, \beta_1 v_1 + \cdots + \beta_{20} v_{20})$, then a single one-dimensional latent variable $z(t) =$ $\beta_1 v_1(t) + \cdots + \beta_{20} v_{20}(t)$ suffices to capture the effects of the qualitative lubricant type. More generally, if the dependence happens to be of the form $y \cong g(x, h_1(v_1, v_2, \ldots), h_2(v_1, v_2, \ldots))$ for some functions $g(\cdot)$, $h_1(\cdot)$ and $h_2(\cdot)$, then a two-dimensional latent variable $z(t) = (h_1(v_1(t), v_2(t), \ldots), h_2(v_1(t), v_2(t), \ldots))$ suffices to capture the effects of the qualitative lubricant type.

The preceding is a rather broad and flexible structure for representing the effects of quantitative variables and qualitative factors on $y$. For even more general $g(\cdot)$ and more complex dependence of the response on $\{v_1, v_2, \ldots\}$, the same arguments behind sufficient dimension reduction (Cook and Ni 2005) imply that the collective effects of $\{v_1, v_2, \ldots\}$ can be represented approximately as a function of the coordinates over some lower-dimensional manifold in the $\{v_1, v_2, \ldots\}$-space. If the manifold is approximately two-dimensional, then $y \cong g(x, h_1(v_1, v_2, \ldots), h_2(v_1, v_2, \ldots))$, and the two-dimensional latent variable representation that we use in our approach will suffice. We illustrate these arguments more concretely with the beam bending example in Section 4.

Section 2 describes our GP model that uses a latent variable representation of qualitative factors, along with the MLE implementation for estimating all covariance parameters, including the latent variable mapping $z(t)$. Section 3 reviews existing GP models for qualitative and quantitative variables and conceptually contrasts them with our proposed method. Section 4 reports numerical comparisons for a number of examples showing that our proposed latent variable method outperforms existing methods on both mathematical and realistic engineering examples.

# 2. LATENT VARIABLE REPRESENTATION OF QUALITATIVE FACTORS

As illustrated with the lubricant example in Section 1, in most physical systems the dependence of the response on the qualitative factors may be represented by a set of underlying quantitative variables. We elaborate on this notion with the beam bending example that will be introduced in Section 4. In light of this, we propose a GP model with a latent variable representation of the qualitative factors. Using the latent variable representation $z(t)$, any standard correlation function for quantitative variables, such as the Gaussian correlation in (2), can be used over the joint space $(x, z)$.

## 2.1. A 1D Latent Variable Representation for $q = 1$

We first describe the approach in the context that we have a single qualitative factor $t$ with $m$ levels (labeled $t = 1, 2, \ldots, m$) and are using a one-dimensional (1D) latent variable $z(t)$ to represent the $m$ levels. The $m$ levels of $t$ will be mapped to $m$ latent numerical values $(z(1), \ldots, z(m))$ for $z$. The input $w = (x, t)$ is therefore mapped to $(x, z(t))$, and using the Gaussian correlation function in (2), our covariance model is

$$R(y(x,t), y(x',t')) = R(y(x, z(t)), y(x', z(t')))$$
$$= \exp\left\{-\sum_{i=1}^{p} \phi_i (x_i - x_i')^2 - (z(t) - z(t'))^2\right\}, \quad (3)$$

where $\phi_i$'s are the correlation parameters for the quantitative variables $x$. Note that there is no correlation parameter for $z$. We take it to be unity, because when $(z(1), \ldots, z(m))$ are estimated in the MLE optimization, their spacing will appropriately account for the correlation between levels of the qualitative factor $t$.

Under the model (3), the log-likelihood function is

$$l(\mu, \sigma, \boldsymbol{\phi}, \mathbf{Z}) = -\frac{n}{2}\ln(2\pi\sigma^2) - \frac{1}{2}\ln|\mathbf{R}(\boldsymbol{\phi}, \mathbf{Z})| - \frac{1}{2\sigma^2}(\mathbf{y} - \mu\mathbf{1})^T \mathbf{R}(\boldsymbol{\phi}, \mathbf{Z})^{-1}(\mathbf{y} - \mu\mathbf{1}), \quad (4)$$

where $n$ is the sample size, $\mathbf{1}$ is an $n$-by-1 vector of ones, $\mathbf{y}$ is the $n$-by-1 vector of observed response values, $\mathbf{Z} = (z(1), \ldots, z(m))$ represents the values of the latent variable corresponding to the $m$ levels of the qualitative variable $t$, and $\mathbf{R}(\boldsymbol{\phi}, \mathbf{Z})$ is the $n$-by-$n$ correlation matrix whose elements are obtained by plugging pairs of the $n$ sample values of $(\boldsymbol{x}, t)$ into (3). Without loss of generality, we set the first level $t = 1$ to correspond to the origin in the latent variable space (i.e., $z(1) = 0$), because in (3) only the relative distances between levels of $t$ in the latent variable space affect the correlation. Fixing $z(1)$ is also necessary to prevent indeterminacy or nonidentifiability during optimization in the MLE algorithm. By "indeterminacy", we mean that the MLE optimization solution would not be unique, since any translation of $\mathbf{Z} = (z(1), \ldots, z(m))$ would give the exact same covariance and likelihood.

In the numerical studies in Section 4, we found that a 1D latent space effectively captures the correlation structure of qualitative factors in a variety of real and realistic examples. However, using the 1D latent representation has certain drawbacks in practice for the following reasons, and we prefer a 2D latent representation. Suppose the qualitative factor $t$ has three levels and that the response correlation $R(y(\boldsymbol{x}, t), y(\boldsymbol{x}, t') | \boldsymbol{\phi})$ for all levels $t$ and $t'$ is the same value (e.g., 0.6). To represent this via (3), the three levels must have equal pairwise distances in the latent variable space, which is impossible using a 1D representation. This is depicted in Figure 1a for the case that $|z(2) - z(1)| = |z(3) - z(2)|$, in which case $|z(3) - z(1)| = 2|z(2) - z(1)|$, so that the correlation between levels $t = 1$ and $t' = 3$ must be smaller than the correlation between the other two pairs of levels. To represent the equal correlation scenario, a two-dimensional (2D) latent space shown in Figure 1b is necessary, in which the three latent mapped values $(z(1), z(2), z(3))$ can form an equilateral triangle. The 2D latent representation also provides correlation structure flexibility in other regards, beyond what the 1D representation can provide.

Another potential issue with a 1D latent representation can occur when the MLE optimizer adjusts the mapped latent values $(z(1), \dots, z(m))$ along the single latent dimension $z$. If any two $z$ values become too close at any point in the optimization, this could cause singularity of the correlation matrix. For example, suppose the initial guesses for two latent points (say $z(1)$ and $z(2)$) are reversed from what their MLEs are. As illustrated in Figure 1c, during the MLE optimization, $z(1)$ and $z(2)$ may need to gradually move toward each other to reverse their positions, which will cause covariance singularity when they get too close. With more qualitative levels, there is a higher probability of encountering singularity during optimization. In contrast, a 2D latent space can reduce the likelihood of singularity significantly, because the points can be moved around more freely in the 2D space. In the 2D space, the positions of $z(1)$ and $z(2)$ can be reversed without ever having to move them too close to each other, as shown in Figure 1d.

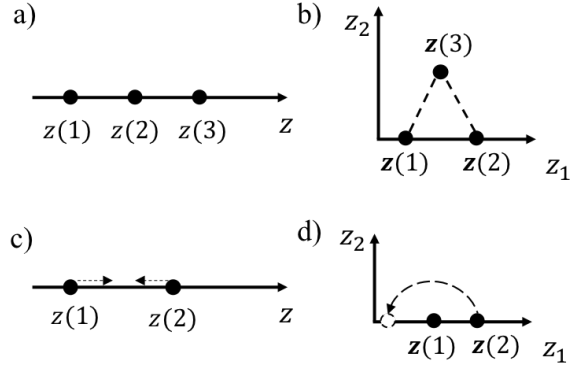

Figure 1: Limitations of 1D latent variable representation: a) in 1D the latent variable mappings cannot represent three equally correlated levels; b) in a 2D latent space, the three latent variable mappings $z(1)$, $z(2)$ and $z(3)$ can be arranged as the vertices of an equilateral triangle to represent equal correlations among all three levels; c) the singularity issue of the covariance matrix when two points become too close to each other when exchanging positions during the MLE optimization search; d) in 2D, the latent variables can move freely to avoid covariance singularity when exchanging positions during the MLE optimization.

## 2.2. A 2D Latent Variable Representation for $q = 1$

As depicted in Figure 1, compared to the single latent variable representation, a 2D latent space provides more flexibility to capture complex correlation structures for qualitative factors, as well as better numerical behavior in the MLE optimization. With a single qualitative factor ($q = 1$), to extend the model (3) to incorporate a two-dimensional latent variable $\mathbf{z} = (z_1, z_2) \in \mathbb{R}^2$, we map the $m$ levels of $t$ to the $m$ points $\{\mathbf{z}(1) = (z_1(1), z_2(1)), \ldots, \mathbf{z}(m) = (z_1(m), z_2(m))\}$ in a 2D latent space. The input $\mathbf{w} = (\mathbf{x}, t)$ maps to $(\mathbf{x}, \mathbf{z}(t))$, and the corresponding Gaussian correlation function is defined as

$$R(y(\mathbf{x}, t), y(\mathbf{x}', t')) = \exp\left\{-\sum_{i=1}^{p} \phi_i (x_i - x_i')^2 - \|\mathbf{z}(t) - \mathbf{z}(t')\|_2^2\right\}, \tag{5}$$

where $\|\cdot\|_2$ denotes the Euclidean 2-norm. The mapped values for the $m$ levels are again estimated via MLE, along with the other covariance parameters. A total of $2(m-1) - 1 = 2m - 3$ separate scalar latent values are required to represent the $m$ different levels of $t$, because (i) similar to the 1D case, the first level of $t$ can always be mapped to the origin (i.e., $\boldsymbol{z}(1) = (0,0)$) to remove the indeterminacy caused by translation invariance; and (ii) to remove indeterminacy due to rotational invariance in the 2D latent space, we can restrict the 2D position of the mapped value $\boldsymbol{z}(2)$ for the second level to lie on the horizontal axis. Figure 2 illustrates this for $m = 3$ by showing three different configurations of three mapped latent values $\{\boldsymbol{z}(1), \boldsymbol{z}(2), \boldsymbol{z}(3)\}$ that are translated and rotated versions of each other. They therefore have the exact same pairwise distances and result in the same covariance structure via (5). Our convention of taking $\boldsymbol{z}(1)$ to be the origin and $\boldsymbol{z}(2)$ to lie on the horizontal axis removes the indeterminacy, as well as reduces the total number of free parameters to estimate from $2m$ to $2m - 3$, which scales linearly with the number of levels of the qualitative variable.

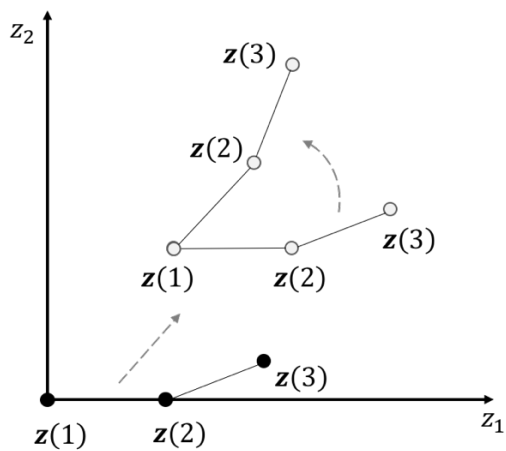

Figure 2: Indeterminancy caused by translation and rotation: Three different configurations for the mapped latent values $\{\boldsymbol{z}(1), \boldsymbol{z}(2), \boldsymbol{z}(3)\}$ having the same pairwise distances and the same covariance structure.

With $m > 3$ levels, one might consider a more general version of our approach that uses an $(m - 1)$-dimensional latent variable representation $\boldsymbol{z} = (z_1, \ldots, z_{m-1}) \in \mathbb{R}^{m-1}$ in (5). Similar to the two-dimensional scenario, to avoid indeterminacy due to rotation/translation invariance, the mapped value for the first level can be taken to be the origin $\boldsymbol{z}(1) = (0, \ldots, 0) \in$ $\mathbb{R}^{m-1}$, and we can likewise restrict $\mathbf{z}(2) = (z_1(2), 0, \ldots, 0) \in \mathbb{R}^{m-1}$, $\mathbf{z}(3) = (z_1(3), z_2(3), 0, \ldots, 0) \in \mathbb{R}^{m-1}$, ..., and $\mathbf{z}(m) = (z_1(m), z_2(m), \ldots, z_{m-1}(m)) \in \mathbb{R}^{m-1}$. Therefore, this model would require estimating $m(m-1)/2$ independent mapped latent variable values in total, which is the same as in the unrestrictive covariance model of Qian et al. (2008). This $m-1$ dimensional latent variable model is a very general covariance structure that allows independent representation of all $m(m-1)/2$ pairwise correlations of the response across the $m$ qualitative levels for $t$.

However, we do not believe such a general $(m-1)$-dimensional latent variable representation is needed for most problems. This is supported by the conceptual arguments given in Section 1, as well as the numerical results in Section 4, for which the correlation structures of qualitative factors have effective low dimensional representation (1D or 2D). A further explanation for this stems from the analogy between our proposed latent variable representation and something akin to multi-dimensional scaling (MDS) (Buja et al. 2007; Kruskal 1964). Given pairwise distances for higher-dimensional data, MDS seeks a lower dimensional representation whose pairwise distances reflect the high-dimensional distances as closely as possible. In many applications, a two-dimensional representation suffices to approximate the high dimensional data (Buja et al. 2007). Analogous to MDS, our proposed approach seeks a two-dimensional latent variable representation whose pairwise distances can approximate that of the full $(m-1)$-dimensional representation of the qualitative factor. One important difference is that the approximation in our latent variable approach is estimated through MLE and takes into account the dependence of the response on the factor levels, which generally enables more extreme dimension reduction than typical MDS applications in which only the inputs variables are considered. This relates closely to the sufficient dimension reduction arguments given in the introduction. Regardless of how many underlying latent physical numerical variables (e.g., $\{v_1, v_2, \ldots, v_{20}\}$ in the lubricant type example in the introduction) truly account for the differences between levels, the functional dependence of the response on these variables will often be such that their effects can be represented by some low-dimensional (2D) latent variable representation. We believe this is the most compelling argument justifying our latent variable approach, although the MDS analogy may perhaps be a helpful way to view the 2D representation.

## 2.3  A 2D Latent Variable Representation with Multiple Qualitative Factors

Now suppose there are $q > 1$ qualitative factors $\boldsymbol{t} = (t_1, t_2, \ldots, t_q)$, where the $j$th factor $t_j \in \{1, 2, \ldots, m_j\}$, and $m_j$ denotes the number of levels of $t_j$. Our approach has a very efficient and natural way to handle multiple qualitative factors that is akin to how multiple numerical input variables are handled in GP modeling. We simply use a different 2D latent variable $\boldsymbol{z}^j$ to represent each qualitative factor $t_j$ ($j = 1, 2, \ldots, q$). As explained earlier, there are $2m_j - 3$ parameters for each $\boldsymbol{z}^j$, so that the total number of parameters is only $\sum_{j=1}^{q}(2m_j - 3)$.

The corresponding Gaussian correlation function for our approach is

$$R\left(y\left(\boldsymbol{x}, \boldsymbol{t} = (t_1, \ldots, t_q)\right), y(\boldsymbol{x}', \boldsymbol{t}' = t'_1, \ldots, t'_q)\right)$$
$$= \exp\left\{-\sum_{j=1}^{p} \phi_j (x_j - x'_j)^2 - \sum_{j=1}^{q} \|\boldsymbol{z}^j(t_j) - \boldsymbol{z}^j(t'_j)\|_2^2\right\}, \qquad (6)$$

where $\boldsymbol{z}^j(l) = (z_1^j(l), z_2^j(l))$ denotes the 2D mapped latent variable for level $l$ of the qualitative factor $t_j$. The $2m_j - 3$ values for the mapped latent variables for each factor $t_j$, along with the parameters $\boldsymbol{\phi}$ and $\mu$ and $\sigma^2$ of the GP model, are estimated via MLE.

In addition to yielding a relatively parsimonious yet flexible parameterization, this approach also has the following desirable characteristic. Using a separate 2D latent variable $z^j$ to represent each qualitative factor results in an approach for handling multiple qualitative factors that is consistent with how numerical input variables are handled in GP modeling. Namely, the numerical inputs ($x_j$'s) and the numerical surrogates ($z^j$'s) for the qualitative factors appear in a similar manner in (6). Moreover, even though we have used the separable (in $x$ and $z$) Gaussian covariance in (6), any covariance model used for numerical variables, including nonseparable ones, can be used over the joint ($x, z$) space.

# 3. COMPARISONS WITH EXISTING APPROACHES FOR QUALITATIVE FACTORS

## 3.1. Unrestrictive Covariance (UC)

A popular approach in the literature for GP modeling with qualitative variables was introduced by Qian et al. (2008) and further developed in Zhou et al. (2011). They assumed

$$R(y(x,t), y(x',t')) = \tau_{t,t'} \exp\{-\sum_{i=1}^{p} \phi_i (x_i - x_i')^2\}, \tag{7}$$

where $\tau_{t,t'}$ is the correlation between the responses corresponding to level $t$ and $t'$. An $m \times m$ correlation matrix $\tau$ with row-$t$, column-$t'$ entry $\tau_{t,t'}$ is used to represent the correlations across all $m$ levels of the qualitative variable. To ensure that the correlation defined in (7) is valid, the matrix $\tau$ must be positive definite with unit diagonal elements (PDUDE). When there are $q > 1$ qualitative factors, one approach is to define a single qualitative factor that represents combinations of levels of all the qualitative factors and then use (7). Alternatively, a somewhat more restrictive structure that was also considered in Qian et al. (2008) is the Kronecker product structure

$$R\left(y\left(\pmb{x},\pmb{t}=(t_1,\dots,t_q)\right),y(\pmb{x}',\pmb{t}'=t_1',\dots,t_q')\right)=\prod_{j=1}^{q}\tau_{t_j,t_j'}^{j}\exp\{-\sum_{i=1}^{p}\phi_i(x_i-x_i')^2\}, \quad (8)$$

where $\tau_{l,l'}^{j}$ represents the correlation between levels $l$ and $l'$ of $t_j$.

Semidefinite programming was used in Qian et al. (2008) for the estimation of $\pmb{\tau}$ to ensure that it is positive definite. Zhou et al. (2011) later simplified the estimation procedure by using hypersphere decompositions (Rebonato and Jäckel 1999). Recently, in Zhang and Notz (2015), the authors showed that it is possible to use indicator variables in the Gaussian correlation function to generate the correlation structure in (8). For positive integers $i$, $l$, and $l'$, define the level indicator functions

$$I_l(i) = \begin{cases} 1 & i = l \\ 0 & i \neq l \end{cases} \quad (9)$$

and

$$W_{l,l'}(i) = \begin{cases} I_l(i) + I_{l'}(i) & if\ l \neq l' \\ I_l(i) & if\ l = l' \end{cases}, \quad (10)$$

and consider the correlation function

$$R\left(y(\pmb{x},\pmb{t}=(t_1,\dots,t_q)),y(\pmb{x}',\pmb{t}'=(t_1',\dots,t_q'))\right)$$
$$= \prod_{j=1}^{q}\exp\left\{-\sum_{l,l'=1}^{m_j-1}\phi_{l,l'}^{j}\left(W_{l,l'}(t_j)-W_{l,l'}(t_j')\right)^2\right\}\exp\{-\sum_{i=1}^{p}\phi_i(x_i-x_i')^2\}, \quad (11)$$

where $\{\phi_{l,l'}^{j}: 1 \leq l, l' \leq m_j - 1\}$ are additional parameters to be estimated via MLE. Zhang and Notz (2015) showed that (11) is equivalent to (8) for $\tau_{l,l'}^{j} > 0$, in that there is a one-to-one correspondence between the $\phi_{l,l'}^{j}$'s (11) and the $\tau_{l,l'}^{j}$'s in (8). Using the formulation in (11) allows one to use standard GP fitting packages to estimate the $\tau_{l,l'}^{j}$'s in the Qian et al. (2008) method with a mild restriction that $\tau_{l,l'}^{j} > 0$. When a single qualitative factor ($q = 1$) is used to represent all the combinations of levels of multiple qualitative factors, (11) reduces to the equivalent version of (7), with $\tau_{l,l'} > 0$. There is no restriction on the elements of $\boldsymbol{\tau}$ as long as it is a PDUDE, so the correlation in (7) is also referred to as an unrestrictive covariance (UC). Because of symmetry, there are $m(m-1)/2$ free parameters to be estimated in $\boldsymbol{\tau}$, which represent all $m(m-1)/2$ pairwise correlations of the qualitative factor levels.

The Qian et al. (2008) method can be viewed as treating the response surfaces for different qualitative levels as separate response functions and using the $\tau_{l,l'}$ parameters to represent the cross-correlation among the different responses in a multivariate GP approach. Our model is fundamentally different in that it considers the response $y$ at different levels of the qualitative factor(s) to be from a single response surface that is continuous over the space of some underlying latent numerical variables that account for the effects of the qualitative factors. As discussed earlier, this is more consistent with how GP models naturally handle quantitative variables, and it also allows any standard GP correlation function to be used over the joint set of quantitative and qualitative variables.

Regarding the latter, the UC model in (7) assumes the correlation structure is separable (i.e., multiplicative) in the qualitative factor $t$ and the numerical variables $x$. In contrast, our proposed latent variable approach is also applicable with correlation functions that are not separable over $(x, t)$, including the non-separable version of the power exponential, Matèrn (Rasmussen et al. 2006), and lifted Brownian (Plumlee and Apley 2017), which provide more flexibility in modeling complex correlations. Another advantage of our model is that the number of parameters $(2m - 3)$ scales linearly with the number of levels of the qualitative variable, which is significantly less than the $m(m-1)/2$ parameters in the UC model when $m$ is large.

### 3.2. Multiplicative Covariance

Qian et al. (2008) also discusses some simplified special cases of the unrestrictive covariance. The simplest model assumes $\tau_{l,l'} = \tau$ for all $l \neq l'$, which is referred to as an exchangeable covariance (EC) (Joseph and Delaney 2007; Qian et al. 2008). Another simplified model termed the multiplicative covariance (MC) (McMillan et al. 1999; Qian et al. 2008) assumes that for all $t \neq t'$

$$\tau_{t,t'} = e^{-(\theta_t + \theta_{t'})}, \tag{12}$$

where $\theta_l > 0$ is the parameter associated with level $l$ of the qualitative factor $t$, and there are $m$ parameters needed in this model. As pointed out in Zhang and Notz (2015), this method is equivalent to using a standard GP for quantitative variables with the qualitative variable represented by the set of indicator variables in (9), analogous to how nominal categorical variables are handled in linear regression.

When $m \leq 3$, the MC model is nearly equivalent to the UC model, with the only difference being that $\tau_{t,t'}$ are restricted to being nonnegative (Zhang and Notz 2015). However, when $m \geq 4$, the MC model has the following undesirable properties, as shown in Zhang and Notz (2015). Suppose $m = 4$ and the response surfaces (over $x$) for levels 1 and 2 are highly correlated, the response surfaces for levels 3 and 4 are highly correlated, but the response surfaces for levels 1 and 2 are very different from the surfaces for levels 3 and 4. According to (12), since each $\theta_l > 0$, in order to make $\tau_{1,2} \approx \tau_{3,4} \approx 1$, we must have $\theta_1, \theta_2, \theta_3$, and $\theta_4$ all close to 0. But in this case the correlation between levels 1 and 3 becomes $\tau_{1,3} = e^{-(\theta_1 + \theta_3)} \approx 1$, which contradicts the assumption that levels 1 and 3 are not correlated. The MC model fails in this case because it uses only $m$ parameters to specify $m(m-1)/2$ pairwise correlations, and the simplified parameterization fails to capture this common physical situation. In contrast, our proposed model can easily handle this case even with the 1D latent variable representations via setting $z(2) \approx 0$ and $z(3) \approx z(4) \gg 0$. We believe that our simplified parameterization using latent variables is more consistent with many physical systems and generally is more effective at capturing commonly occurring correlation structures, while requiring only a small number of parameters.

### 3.3. Additive GP Model with Qualitative Variables

The UC and MC models both assume multiplicative forms of correlations across the quantitative factors and qualitative factors. Deng et al. (2017) pointed out that a potential drawback of (8) is that if some $\tau^j_{t_j, t'_j}$ is zero, then the response correlation at levels $t_j$ and $t'_j$ for factor $j$ must be zero for all level combinations of the other factors and for all $x$. Instead of using a multiplicative structure, Deng et al. (2017) proposed the additive covariance structure

$$K(y(x,t), y(x',t')) = \sum_1^q \sigma_j^2 \tau^j_{t_j, t'_j} R(x, x'|\phi^{(j)}), \qquad (13)$$

where $R(x, x'|\phi^{(j)})$ is the Gaussian correlation function defined in (2) with correlation parameters $\phi^{(j)}$ associated qualitative factor $t_j$, $\sigma_j^2$ is a prior variance term associated with qualitative factor $t_j$, and $\tau^j_{t_j, t'_j}$ has the same definition as in (8). This covariance model is equivalent to assuming

$$y(x, t_1, \ldots, t_q) = \mu + G_1(x, t_1) + \cdots + G_q(x, t_q),$$

where $\mu$ is the overall mean, and the $G_j$'s are independent zero-mean GPs, each with covariance functions over $(x, t_j)$ given by the individual terms in (13). This formulation allows a different covariance structure over $x$ for each qualitative factor. When there is only one qualitative factor, this model is equivalent to the covariance model (8). For $q > 1$, Deng et al. (2017) argues that it provides more flexibility for modeling complex computer simulations than the model (8), which assumes a fixed covariance structure over $x$ for all quantitative factors.

It should be noted that if all categorical inputs have two levels, our LVGP covariance (6), the Qian et al (2008) Kronecker product covariance (8), and the MC covariance (12) are equivalent to the standard GP approach for numerical inputs using binary numerical coding for the two-level categorical inputs. Hence, we focus on the situation of more than two levels for the categorical inputs.

## 4. NUMERICAL COMPARISONS

In this section, we conduct numerical studies to investigate the effectiveness of the proposed latent variable model (6) with Gaussian correlation function. We compare the proposed method with the three covariance structures reviewed in the previous section:

(a) Unrestrictive correlation (UC) defined in (8) (Qian et al., 2008; Zhou et al., 2011), using the equivalent reformulation in (11) discussed in Zhang and Notz (2015);

(b) Multiplicative correlation (MC) defined in (12) (McMillan et al. 1999; Qian et al. 2008; Zhang and Notz 2015), using the equivalent reformulation with indicator variables (9) discussed in Zhang and Notz (2015);

(c) Additive GP with unrestrictive correlation (Add_UC), defined in (13), which is equivalent to UC when there is only a single qualitative factor.

The Gaussian correlation function in (2) is used for all quantitative variables $x$ in these four methods. To evaluate the model accuracy of each method, we use the relative root mean squared error (RRMSE) for the fitted GP model predictions over $N = 10,000$ hold-out test points:

$$RRMSE = \sqrt{\frac{\sum_{i=1}^{N}(\hat{y}(\boldsymbol{w}_i) - y(\boldsymbol{w}_i))^2}{\sum_{i=1}^{N}(y(\boldsymbol{w}_i) - \bar{y})^2}}, \quad (14)$$

where $\hat{y}(\boldsymbol{w}_i)$ and $y(\boldsymbol{w}_i)$ denote the predicted and the true values at input test location $\boldsymbol{w}_i$ respectively, and $\bar{y}$ is the average of the true responses at the 10,000 test points. The 10,000 test points are generated uniformly for both the quantitative variables and qualitative factors. For each example, we used 30 replicates, where on each replicate we generated a different "training" design, the data from which were used to fit the four covariance models, and then we calculated the resulting RRMSE for the four models. Each training design was a maximin Latin hypercube design (LHD), and the design sizes were chosen so that the RRMSE for the best model for each example was less than 0.1, to ensure that the designs were of sufficient size to allow reasonable prediction accuracy.

We fit the UC and MC models through the same optimization routine for MLE used in our latent variable model. MATLAB code from the supplemental materials of Deng et al. (2017) was used to fit the Add_UC model. To have a common basis for comparison, when fitting all models, we used 200 random initial guesses for the GP hyper-parameters to help ensure good MLE solutions. During optimization, the correlation parameters for all quantitative inputs are reparametrized as $\theta_i = log_{10}(\phi_i)$, with $\theta_i \in [-3,3]$, and each latent variable is restricted to interval $z_i^j(l) \in [-2,2]$. Formerly, in general, we had used a much larger interval $[-10,10]$ over which to search for the MLEs of the latent variables. However, their MLEs were almost always much smaller than this, so we now restrict the search range to $[-2,2]$. This typically allows sufficiently small correlations between levels, when small correlations are needed. A LHD is used for generating the 200 random initial guesses to cover the search space as evenly as possible. For the MLE optimization, we use the MATLAB function *fmincon*, which uses an interior-point method with BFGS for a Hessian approximation.

## 4.1. Mathematical Examples

We first test the four methods on two mathematical functions that have been used in the literature as benchmark problems involving qualitative factors (Deng et al. 2017; Swiler et al. 2014). We also include four engineering examples that are popular choices for assessing surrogate models with numerical inputs. In these examples, we converted some of the numerical input variables to qualitative factors. This has the benefit of providing a second means of assessing the effectiveness of our covariance model. Namely, since the qualitative factors in this case are truly due to some underlying numerical variables, and we know what values of the numerical variable correspond to the factor levels, we can compare the true values with our estimated latent variable values.

**Math Function 1**

The first mathematical test function adapted from Swiler et al. (2014) has one qualitative variable $t$ with five levels, and two continuous variables $x_1, x_2 \in [0,1]$. This function has regions where the response behaviors at different qualitative levels are very similar. The different levels of $t$ are associated with the coefficients of the second term in the following definition of the function

$$y(\boldsymbol{x}, t) = \begin{cases} 7\sin(2\pi x_1 - \pi) + \sin(2\pi x_2 - \pi) & if\ t = 1 \\ 7\sin(2\pi x_1 - \pi) + 13\sin(2\pi x_2 - \pi) & if\ t = 2 \\ 7\sin(2\pi x_1 - \pi) + 1.5\sin(2\pi x_2 - \pi) & if\ t = 3 \\ 7\sin(2\pi x_1 - \pi) + 9.0\sin(2\pi x_2 - \pi) & if\ t = 4 \\ 7\sin(2\pi x_1 - \pi) + 4.5\sin(2\pi x_2 - \pi) & if\ t = 5 \end{cases} \quad (15)$$

The levels of $t$ therefore have a true ordering 1-3-5-4-2, which follows by comparing the coefficients of the third terms in (15). To fit the four GP models, a different training set of size 70 is generated for each of the 30 replicates, using maximin LHDs for the quantitative variables $x_1$ and $x_2$ with the levels of $t$ randomly assigned. The left panel of Figure 3 shows the prediction accuracy over the 10,000 hold-out points via boxplots of the 30 RRMSE values across the 30 replicates. The median values of the RRMSE for the ADD_UC, UC, MC, and LV models are 0.181, 0.103, 0.134, and 0.015, respectively. Thus, our LV model achieved an average RMSE that was roughly an order of magnitude smaller than for the other models for this example.

In addition to more accurate response predictions, our LV model can provide valuable insight into the effects of the qualitative factor on the response. For example, the left panel of Figure 4 displays the estimated 2D latent variables for a typical replicate for this example. Even though the 2D latent variables were estimated, their MLE values fell almost exactly on a straight line. The straight line corresponds to the $z_1$ axis, since $\boldsymbol{z}(1)$ is restricted to the origin, and $\boldsymbol{z}(2)$ is restricted to falling on the $z_1$ axis. This result is desirable, since the qualitative variable truly corresponds to a 1D latent variable in this example, as discussed above. Moreover, the estimated latent variables are correctly ordered as 1-3-5-4-2 with levels 1 and 3 positioned very close to each other. This is very consistent with (15), in which the response surfaces at level 1 and 3 have the smallest differences. The response surface at level 2 has the most substantial differences with the surface at level 1, and our estimated $\boldsymbol{z}(2)$ is correctly positioned as the farthest from the origin. It should be noted that the estimated latent variables do have non-zero values in the $z_2$ coordinate, although they are so small that they are visually indiscernible in Figure 4.

**Math Function 2**

The second function used for comparing the four models is from Deng et al. (2017) with $p = 5$ quantitative variables and $q = 5$ qualitative factors, each having three levels:

$$y = \sum_{i=1}^{5} \frac{x_i(t_{6-i}-2)}{80} + \prod_{i=1}^{5} \cos\left(\frac{x_i}{\sqrt{i}}\right) \sin\left(\frac{50(t_{6-i}-2)}{\sqrt{i}}\right), \tag{16}$$

where $-100 \leq x_i \leq 100$, for $i = 1, \ldots, p$, and $t_j$ ($j = 1, \ldots, 5$) are the five qualitative factors, each having three levels $\{1, 2, 3\}$. We generated a maximin LHD of size $n = 100$ as the training set for the $x_i$'s in each of the 30 replicates, with the levels of the $t_j$'s randomly assigned. The qualitative factors affect the response $y$ in a nearly additive manner, and as a result, the additive GP model outperforms the UC and MC models, as seen in the middle boxplots in Figure 3. However, our proposed model works even better than the additive GP model, having the lowest median RRMSE value of 0.045, which is roughly four times smaller than the additive GP model's 0.185 RRMSE. The estimated latent variables associated with $t_1$ are shown in Figure 4 for a typical replicate, for which the three levels are approximately equally spaced along the $z_1$ axis and correctly ordered as 1-2-3, which again agrees very closely with the true underlying numerical $t_j$ in (16). The estimated latent variables for $t_2$ to $t_6$ were in similar agreement and are not shown in Figure 4.

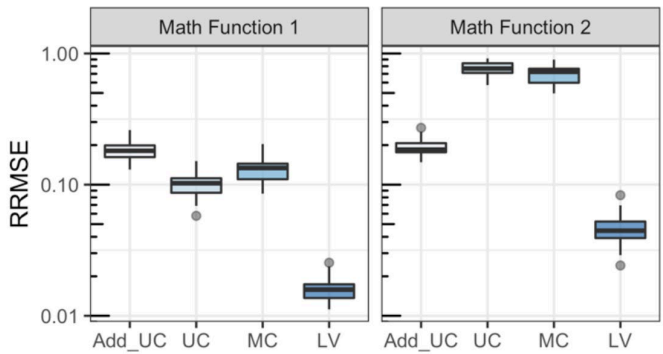

Figure 3: Boxplots of RRMSE over 30 replicates for the two mathematical functions with design sizes of $n = 70$ and 100, respectively. Our LV approach outperforms the other three methods. Note that the y-axis is in log scale.

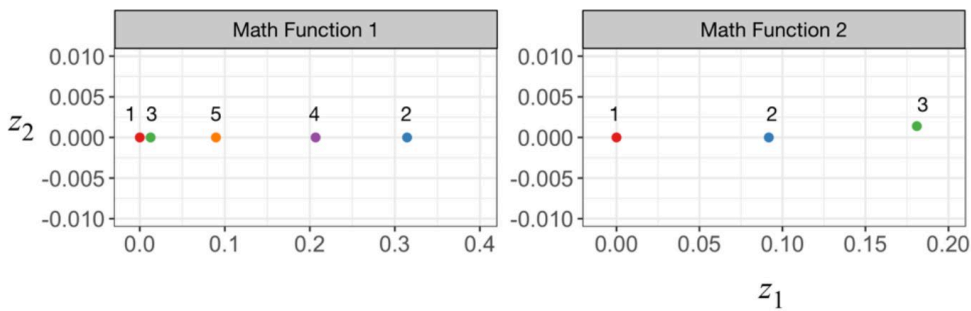

Figure 4: Estimated 2D latent variables $\mathbf{z} = (z_1, z_2)$ representing the levels of the qualitative factors in the three mathematical test functions for typical replicates: the values of $z_2$ are nearly zero for both examples, indicating the true 1D latent structure was correctly identified; and the estimated spacing of between the latent values match with the settings in each example.

### 4.2. Engineering Examples

We now compare the approaches on a set of real engineering examples. The response surface in each example has an explicit mathematical form but is based on the physics of the problem. In all examples, training sets are generated through maximin LHDs for quantitative variables, and the levels of qualitative factors are randomly sampled.

**Beam Bending**

First consider the classic beam bending problem in which the qualitative factor is the cross-sectional shape of the beam with six levels: circular, square, I-shape, hollow square, hollow circular, and H-shape (shown in Figure 5). The beam has an elastic modulus E = 600GPa, and is operating within its linear elastic range. The beam is fixed on one end, and a force of P = 600N is applied vertically at the free end. The response $y$ is the amount of deformation at the free end. In addition to the cross-sectional shape represented by a qualitative factor $t$, there are two numerical input variables: beam length $L$ and beam width (which is the same as beam height) $h$.

This is a good example to illustrate the point we made in the introduction about the strong physical justification for our approach. Namely, in computer simulation models, any qualitative factor must always affect the response only via some set of underlying numerical variables $\{v_1(t), v_2(t), v_3(t), ...\}$. In a finite element simulation of the beam bending example, $\{v_1(t), v_2(t), v_3(t), ...\}$ would be the complete geometric positions (normalized by the cross-sectional "size" parameter $h$) of all elements in the finite element mesh of the beam cross-section for section type $t$. However, the physics of this beam bending problem are transparent enough that we know the beam deflection $y$ depends on $\{v_1(t), v_2(t), v_3(t), ...\}$ via the response function

$$y(L, h, t) = \frac{L^3}{3 \times 10^9 h^4 I},$$

where $I = I(t) = I(v_1(t), v_2(t), v_3(t), ...)$ is the normalized (by $h$) moment of inertia of the cross-section, which is a function of the complete high-dimensional geometric descriptors $\{v_1(t), v_2(t), v_3(t), ...\}$ of the cross-section. Consequently, the underlying high-dimensional variables that govern the effect of the qualitative factor $t$ on $y$ can be mapped down to a single numerical variable $I(t)$. When applying our LV approach below, we do not incorporate this knowledge and, instead, let the approach attempt to discover the underlying effect of the qualitative cross-sectional shape factor.

In light of the preceding, if our LV approach performs effectively, the estimated latent variable mapping $\mathbf{z}(t)$ will represent the normalized moment of inertia $I(t)$. From basic mechanics, the normalized moments of inertia for the six cross-sections in Figure 5 are $I_1 = \pi/64 = 0.0491$, $I_2 = 1/12 = 0.0833$, $I_3 = 0.0449$, $I_4 = 0.0633$, $I_5 = 0.0373$, and $I_6 = 0.0167$, and their inverses (which turn out to be very closely-related to the mapping $\mathbf{z}(t)$) are $1/I_1 = 20.4$, $1/I_2 = 12.0$, $1/I_3 = 22.3$, $1/I_4 = 15.8$, $1/I_5 = 26.8$, and $1/I_6 = 59.9$. Notice that the H-shaped cross-section (level $t = 6$) has substantially different $1/I$ than the other cross sections, and the six cross-sections are ordered from largest to smallest $1/I$ according to levels 6, 5, 3, 1, 4, then 2. The ordering and relative spacing agrees nearly perfectly with the estimated latent variables $\mathbf{z}(1)$—$\mathbf{z}(6)$ in Figure 7. Consequently, our LV model correctly discovered the underlying mapped latent variable $\mathbf{z}(t)$ that dictates the effect of the qualitative factor on $y$.

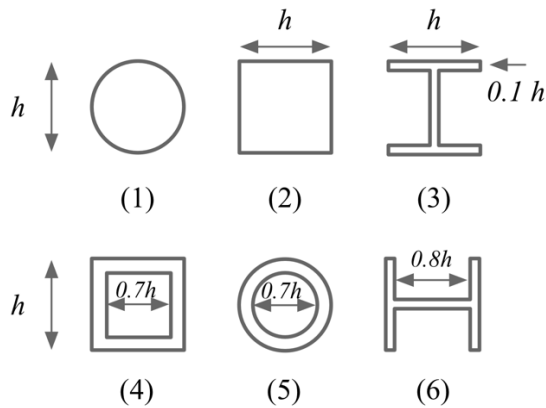

Figure 5: Six cross-sectional shapes, representing six levels of the qualitative factor for the beam bending example: (1) circular cross-section with diameter $h$; (2) square cross-section with height and width $h$; (3) I-shaped cross-section with height and width $h$ and thickness $0.1h$; (4) hollow square cross-section with outer side length $h$ and thickness $0.15h$; (5) hollow circular cross-section with outer diameter $h$ and thickness $0.15h$; (6) H-shape cross-section with height and width h and thickness $0.1h$.

**Borehole**

A commonly used function to study computer simulation surrogate modeling is the borehole function,

$$y = 2\pi T_u (H_u - H_l) \left\{ \log\left(\frac{r}{r_w}\right) \left( 1 + 2 \frac{L T_u}{\log\left(\frac{r}{r_w}\right) r_w^2 K_w} + \frac{T_u}{T_l} \right) \right\}^{-1},$$

where the 8 inputs are $(T_u, r, r_w, H_u, T_l, H_l, L, K_w)$. See Morris et al. (1993) for a full description of the variables. We treat $r_w$ and $H_l$ as qualitative factors with three and four levels, respectively. The levels for this and the subsequent examples are listed in Table 2.

**OTL**

The midpoint voltage of a transformerless (OTL) circuit function is

$$y = B \frac{(V_{b1} + 0.74)(R_{c2} + 9)}{B(R_{c2} + 9) + R_f} + 11.35 \frac{R_f}{B(R_{c2} + 9) + R_f} + 0.74 B \frac{R_f}{R_{c1}} \frac{R_{c2} + 9}{B(R_{c2} + 9) + R_f},$$

where $V_{b1} = 12 R_{b2} / (R_{b1} + R_{b2})$, and the inputs are $(R_{b1}, R_{b2}, R_f, R_{c1}, R_{c2}, B)$. See Ben-Ari and Steinberg (2007) for details. We treat $R_f$ and $B$ as qualitative factors having 4 and 6 levels, respectively.

**Piston**

The last example models the cycle time for a piston moving within a cylinder as

$$y = 2\pi \sqrt{\frac{M}{k + S^2 \frac{P_0 V_0 T_a}{V^2 T_0}}}, \text{ where } V = \frac{S}{2k} \sqrt{A^2 + 4k \frac{P_0}{T_0} T} \text{ and } A = P_0 S + 19.62 M - \frac{k V_0}{S}.$$

The inputs are $(M, S, V_0, k, P_0, T_a, T_0)$. See Sacks et al. (1989) for details. We treat the two variables $P_0$ and $k$ as qualitative factors each having 3 and 5 levels, respectively.

Table 1: Quantitative input ranges for the four engineering examples

| Bending | Borehole | OTL circuit | Piston |
|---|---|---|---|
| $L \in [10, 20]$ | $R_{b1} \in [50, 150]$ | $R_{b1} \in [50, 150]$ | $M \in [30, 60]$ |
| $h \in [1, 2]$ | $R_{b2} \in [25, 70]$ | $R_{b2} \in [25, 70]$ | $S \in [0.005, 0.020]$ |
| | $R_{cf} \in [1.2, 2.5]$ | $R_{cf} \in [1.2, 2.5]$ | $V_0 \in [0.002, 0.010]$ |
| | $R_{c2} \in [0.25, 1.20]$ | $R_{c2} \in [0.25, 1.20]$ | $T_a \in [290, 296]$ |
| | | | $T_0 \in [340, 360]$ |

Table 2: Qualitative factors and their levels for the four engineering examples

| Level | Bending $t$ | Borehole $t_1 = r_w$ | Borehole $t_2 = H_l$ | OTL circuit $t_1 = R_f$ | OTL circuit $t_2 = B$ | Piston $t_1 = P_0$ | Piston $t_2 = k$ |
|---|---|---|---|---|---|---|---|
| 1 | Circular | 0.05 | 700 | 0.5 | 50 | 9000 | 1000 |
| 2 | Square | 0.1 | 740 | 1.2 | 100 | 10000 | 2000 |
| 3 | I-shape | 0.15 | 780 | 2.1 | 150 | 11000 | 3000 |
| 4 | Hollow Circular | | 820 | 2.9 | 200 | | 4000 |
| 5 | Hollow Square | | | | 250 | | 5000 |
| 6 | H-shape | | | | 300 | | |

Table 1 lists the ranges of all quantitative input variables in these four examples, and Table 2 displays the qualitative factors and their levels. Figure 6 shows boxplots of the RRMSE over the 10,000 hold-out test prediction points across 30 replicates. On each replicate, a different maximin LHD was generated and each of the four models refit. Our LV model performed the best in terms of having the lowest RRMSE across all four examples. The MC and UC models had similar performance, except that MC worked a little better than UC on the OTL example. The Add_UC model had the highest error across all four examples, which might be because these real engineering examples do not have the additive structure that it assumes. Figure 7 displays the estimated 2D latent variables from our method for the four examples, from which we see that most are positioned nearly exactly along the horizontal $z_1$ axis, correctly indicating that there is a single latent numerical variable associated with each qualitative factor $t_j$ and matching the settings in Table 2. In addition, the ordering and relative distances between the numerical values of the mapped qualitative levels in Figure 7 closely mimic those for the true levels in Table 2. Recall that for the bending example, the inverse moments of inertia for the six cross-sectional shapes are $1/I_6 = 59.9$, $1/I_5 = 26.8$, $1/I_3 = 22.3$, $1/I_1 = 20.4$, $1/I_4 = 15.8$, and $1/I_2 = 12.0$, which agrees nearly perfectly with what is shown in Figure 7.

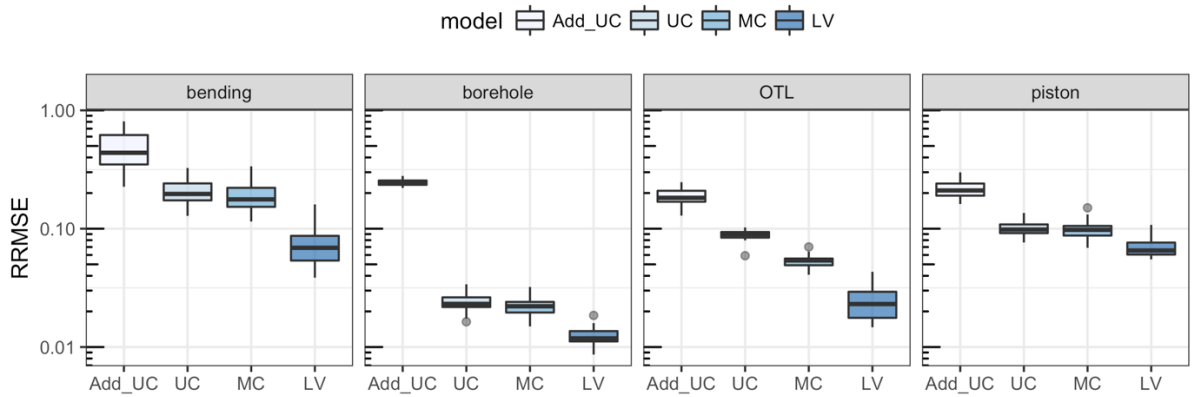

Figure 6: boxplots of RRMSE across 30 replicates for the four engineering examples with $n = 60, 80, 60$, and $100$, respectively. Our LV model achieves the smallest RRMSE for each example. Note that the y-axis is in log scale.

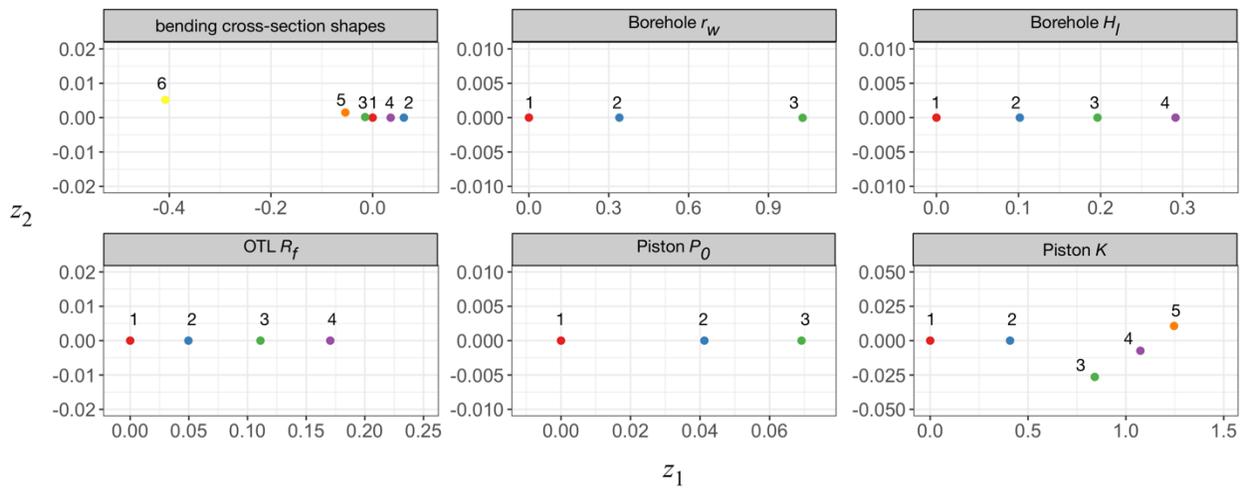

Figure 7: Estimated 2D latent variables $\mathbf{z} = (z_1, z_2)$ representing the levels of the qualitative factors in the four engineering examples for a typical replicate: the values of $z_2$ are small compared to $z_1$, indicating that the estimated latent representation is a one-dimensional representation that closely matches the settings in Table 2.

## 4.3. Borehole Example Revisited with A True Latent Space That is 2D

In all of the preceding examples, the effects of each qualitative factor could be reduced to a function of a single underlying numerical variable, so that the true latent numerical space for each qualitative factor was 1D. In this section we modify the borehole example by creating a qualitative factor $t$ having 12 levels that represent 12 discrete combinations of the two underlying numerical variables $r_w$ and $H_l$. The mapping from $(r_w, H_l)$ to the level of $t$ is listed in Table 3. The other quantitative input variables all have the same ranges shown in Table 1. This example represents the case where multiple underlying numerical variables vary across the levels of a qualitative factor, and we demonstrate below that our LV model can successfully reveal the underlying structure.

Table 3: Mapping from the 2D underlying numerical variables $(r_w, H_l)$ to the single qualitative factor t in the revised borehole example

| Level of $t$ | $r_w$ | $H_l$ | Level of $t$ | $r_w$ | $H_l$ | Level of $t$ | $r_w$ | $H_l$ |
|---|---|---|---|---|---|---|---|---|
| 1 | 0.05 | 700 | 5 | 0.10 | 700 | 9 | 0.15 | 700 |
| 2 | 0.05 | 740 | 6 | 0.10 | 740 | 10 | 0.15 | 740 |
| 3 | 0.05 | 780 | 7 | 0.10 | 780 | 11 | 0.15 | 780 |
| 4 | 0.05 | 820 | 8 | 0.10 | 820 | 12 | 0.15 | 820 |

Figure 8a plots the estimated 2D latent variables associated with the qualitative factor $t$, from which we see that the 12 levels of $t$ are arranged into three groups, each representing a different level of $r_w$. Moreover, within each group, as $H_l$ increases, the points move predominantly along the $z_1$ direction. Thus, the estimated 2D latent variables have successfully revealed the dependence of the qualitative factor on the underlying numerical variables $r_w$ and $H_l$, with $z_2$ approximately representing $r_w$, and $z_1$ approximately representing a combination of $H_l$ and $r_w$.

Notice that the levels of $r_w$ and $H_l$ are evenly spaced in their original units as shown in Table 3, but the estimated $z_1$ and $z_2$ values are not evenly spaced in Figure 8a. The reason is that in our LV model, the distances between latent variables depends on the response correlation across the qualitative levels, which depends not only on the distances between the underlying inputs but also on the behavior of the response. The response surface contour plot in Figure 8b further illustrates the reason: when $r_w$ is at its lower level 0.05, the response does not change as much along the $H_l$ dimension as when $r_w$ is at its higher level 0.15. Consequently levels 1—4 are more closely spaced in Figure 8a than are levels 9—12. In this sense, our LV model has correctly identified the structural dependence of the qualitative levels on a set of underlying numerical variables, in terms of capturing the response similarities/differences across the levels of the factor.

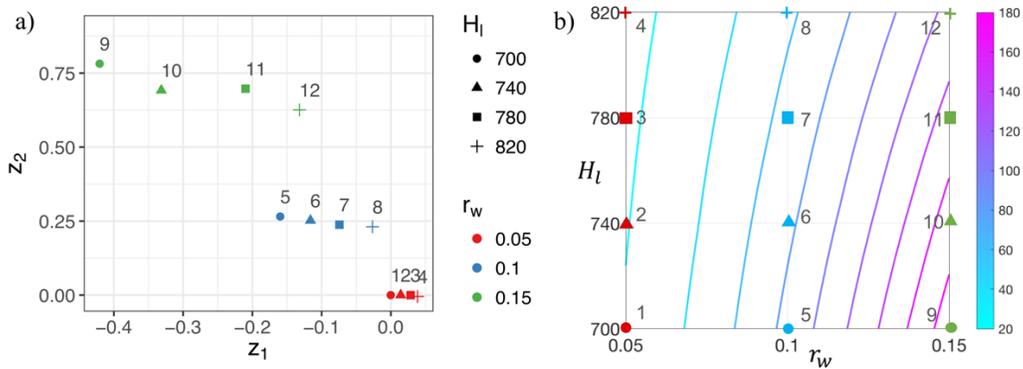

Figure 8: a) estimated 2D mapped latent variables representing the 12 levels of the qualitative factor t in the revised borehole example. The latent representation successfully uncovered the structural dependence of the factor levels on the two underlying numerical variables: the three levels of $r_w$ (represented by colors) are distributed along $z_2$ dimension and within each $r_w$ group the four levels of $H_l$ correspond to $z_1$ varying; b) contour plot of the response in revised borehole example as a function of $r_w$ and $H_l$ with the other numerical variables in Table 1 fixed their mean values.

# 5. WHY USE QUALITATIVE FACTORS AT ALL?

Aside from the superior numerical performance for the LVGP approach demonstrated in our examples, its main justification is the observation that the effects of qualitative factors on a numerical response must always be due to differences in some set of underlying numerical variables $\{v_1(t), v_2(t), v_3(t), ...\}$ across the different levels of the factor. In light of this, one may question whether it would be better to simply identify what are the variables $\{v_1(t), v_2(t), v_3(t), ...\}$ that vary across the levels $t$ of the factor, and then to include these as numerical inputs in a standard GP model for only numerical variables. Identifying the numerical variables should generally be straightforward, albeit perhaps tedious, since whoever coded the simulation must know which variables he/she included in the code. Hence, the appropriateness of a purely numerical GP model largely depends on the dimension of $\{v_1(t), v_2(t), v_3(t), ...\}$ and on the level of prior knowledge regarding how they collectively impact the response variable.

To illustrate, reconsider the beam bending example, in which $\{v_1(t), v_2(t), v_3(t), ...\}$ for the cross-sectional qualitative factor are the complete set of 2D coordinates for every integration point in the finite element mesh of the cross-section. If 1,000 integration points are used, then there are 2,000 underlying numerical variables $\{v_1(t), v_2(t), v_3(t), ..., v_{2000}(t)\}$ that vary as the level $t$ (cross-section shape) varies. The best way to handle this is to have prior knowledge of the physics of the system and to know in advance that $\{v_1(t), v_2(t), v_3(t), ..., v_{2000}(t)\}$ only impact the response via the one-dimensional function $I(t) = I(v_1(t), v_2(t), v_3(t), ..., v_{2000}(t))$ that is the moment of inertia. In this case, it would be naive to treat the cross-sectional shape as a qualitative factor, as opposed to representing cross-sectional shape via the single numerical variable $I(t)$ and including it in a standard GP model for numerical-only inputs (although our example below indicates that one might not lose too much if the LVGP approach is used).

Such strong prior knowledge of how $\{v_1(t), v_2(t), v_3(t), ...\}$ impact the response is not generally available. If $\{v_1(t), v_2(t), v_3(t), ...\}$ is low-dimensional (e.g., only one or two variables), then one should probably forego a qualitative factor treatment and, instead, include $\{v_1, v_2, v_3, ...\}$ as additional numerical variables in the simulation experiments. This would entail varying $\{v_1, v_2, v_3, ...\}$ over some experiment designed for numerical variables, conducting the simulation runs at these values, and then using a GP model for numerical variables to model the response surface.

On the other hand, if $\{v_1(t), v_2(t), v_3(t), ...\}$ is high-dimensional, it may be impossible to include them all as additional numerical variables in the simulation. This is clearly the case for the beam bending example, for which one would never attempt to include $\{v_1, v_2, v_3, ..., v_{2000}\}$ as 2,000 additional numerical variables in the simulation experiment and in the GP surrogate model. Instead, with only six different levels, one would be far better off treating the cross-section as qualitative factor and using the LVGP approach.

An added benefit of the LVGP approach is that it can help to discover the low-dimensional latent variables $\{z_1(t), z_2(t)\} = \{z_1(v_1(t), v_2(t), v_3(t), ...), z_2(v_1(t), v_2(t), v_3(t), ...)\}$ that capture the effects of the underlying high-dimensional variables $\{v_1(t), v_2(t), v_3(t), ...\}$ on the response. This is illustrated in the top-left panel of Fig. 7, which shows that there is only a single low-dimensional combination $z_1(t)$ (since the estimated 2D $\mathbf{z}(t)$ values fall on nearly a straight line) that captures the impact of $\{v_1(t), v_2(t), v_3(t), ..., v_{2000}(t)\}$ on the beam deflection. Recall that this learned $z_1(t)$ corresponds very closely to the inverse of $I(t)$.

Similarly, for the example in Fig. 8, the LVGP approach has learned the correct two-dimensional latent structure of how the qualitative factor affects the response.

In general, it may be better to forego a qualitative factor GP model and instead represent them as numerical inputs in a standard GP model if either (*i*) there are only a few underlying numerical variables that differ across the levels of the qualitative factor or (*ii*) there are many underlying numerical variables, but one has strong prior knowledge that they collectively effect the response only via a few low-dimensional combinations, and the functional forms of these combinations are known. If many underlying numerical variables differ across levels, and one does not understand the physics clearly enough to identify a few low-dimensional combinations on which the response depends, then a GP model for qualitative inputs should be used.

The remainder of this section investigates three additional examples related to the above points. The first two are modifications of the earlier simulation examples. In the modified versions, we compare our LVGP approach (in which only the qualitative levels of the input are available) with an approach that treats the underlying numerical variables $\{v_1(t), v_2(t), v_3(t), ...\}$ as available and uses them in a standard GP model for numerical inputs. We refer to the latter as the benchmark numerical GP (BNGP) approach, since it uses information (the underlying numerical variables) that is not used in the LVGP approach or the other approaches for qualitative inputs. The same designed experiments are used for both methods, so that the numerical variables used in the BNGP approach are only evaluated at locations corresponding to the qualitative levels.

***LVGP vs. BNGP, and the Impact of Dimension.*** In the first example, we replace the qualitative variable $t$ in Math Function 1 described in (15) by a set of $J$ underlying numerical variables $\{v_1(t), v_2(t), v_3(t), ..., v_J(t)\}$ for various $J$. Specifically, the response is

$$y\left(\mathbf{x}, v_1(t), v_2(t), ... v_J(t)\right) = 7\sin(2\pi x_1 - \pi) + \left[J^{-1/2} \sum_{j=1}^{J} v_j(t)\right] * \sin(2\pi x_2 - \pi). \quad (17)$$

To be consistent with the $y(\mathbf{x}, t)$ response surface in (15), we chose the $5 \times J$ values for $\{v_1(t), v_2(t), ... v_J(t): t = 1, 2, ..., 5\}$ so that $\{J^{-1/2} \sum_{j=1}^{J} v_j(t) : t = 1, 2, ..., 5\} = \{1, 13, 1.5, 9.0, 4.5\}$. Beyond that, we randomly generated the values for $\{v_1(t), v_2(t), ... v_J(t): t = 1, 2, ..., 5\}$. That is, we used the basis vectors $\mathbf{A} = [\mathbf{a}_1, \mathbf{a}_2, ..., \mathbf{a}_J]$ for the $J$-dimensional space, where $\mathbf{a}_1 = J^{-1/2} \mathbf{1}$, and $\mathbf{a}_j = J^{-1/2}(J-1)^{-1/2}(J\mathbf{e}_j - \mathbf{1})$ for $j = 2, 3, ..., J$, with $\mathbf{1}$ and $\mathbf{e}_j$ denoting the $J$-length column vector of ones and the $J$-length column vector of zeros with a one in the $j$-th position, respectively. Then, for each $t = 1, 2, ..., 5$, we used $[v_1(t), v_2(t), ... v_J(t)]^T = \mathbf{A}[v(t), u_2(t), ... u_J(t)]^T$ with $\{v(t): t = 1, 2, ..., 5\} = \{1, 13, 1.5, 9.0, 4.5\}$ and the $5 \times (J-1)$ values for $\{u_2(t), u_3(t), ... u_J(t): t = 1, 2, ..., 5\}$ randomly generated from a uniform distribution over the interval $[0, 10]$.

We conducted 20 replicates of the example, where on each replicate we generated a different set of $5 \times (J-1)$ uniform random numbers for $\{u_2(t), u_3(t), ... u_J(t): t = 1, 2, ..., 5\}$ and a different experimental design. For the latter, we generated a size-$n$ LHD in the $\{x_1, x_2\}$ space and then assigned the level for $t$ for each of the $n$ run by randomly sampling one of its five levels. The BNGP model was fit to the same data as the LVGP model but using the underlying numerical $\{v_1(t), v_2(t), ... v_J(t)\}$ instead of $t$. Figure 9 compares the RRMSEs across 20 replicates for five different DOE sizes ($n = 30, 40, 50, 60$, and $70$) and for four different values of $J$ (1, 3, 5, 10).

The main conclusion drawn from Figure 9 is that if the underlying numerical variables are low-dimensional ($J = 1$), very little accuracy is lost if we use the LVGP approach, relative to using the BNGP approach that incorporates the numerical variable information; and if the underlying numerical variables are higher-dimensional ($J \geq 3$), the LVGP gives much better accuracy than the BNGP approach. We note that for $J = 1$ and the smaller designs ($n < 50$ roughly), the BNGP approach does indeed perform slightly better than the LVGP approach, but the difference becomes negligible for the larger designs ($n > 50$ roughly).

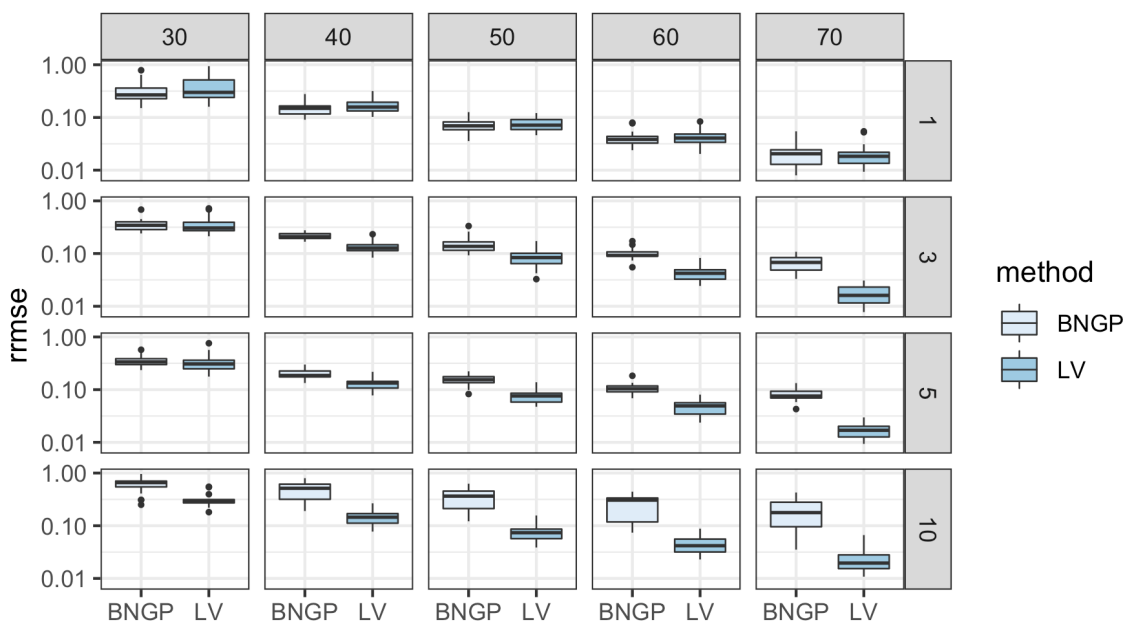

Figure 9: RRMSE comparison (each boxplot is for 20 replicates) of BNGP vs. LVGP for the example in (17) with design sizes $n = 30, 40, 50, 60,$ and $70$ (corresponding to columns) when the dimension of the underlying numerical variables is $J = 1, 3, 5, 10$ (corresponding to rows). Our LVGP model has only slightly higher error than the benchmark BNGP model that uses the underlying numerical $v(t)$ when $J = 1$, and the differences decrease with larger $n$. The BNGP model degrades significantly when the dimension of the underlying numerical variables increases.

***Example with a 10-D $\{v_1(t), v_2(t), v_3(t), ...\}$ that Cannot be Reduced to a 2-D $z(t)$.***

Throughout, we have used a two-dimensional latent variable $z(t)$ for each qualitative factor,

based in part on sufficient dimension reduction arguments that a two-dimensional $\mathbf{z}(t)$ should provide a reasonable approximation of the effects of the underlying $\{v_1(t), v_2(t), v_3(t), ...\}$ for many qualitative factors. This was illustrated with the beam bending example, in which $\{v_1(t), v_2(t), v_3(t), ..., v_{2000}(t)\}$ is very high-dimensional, but their collective effect on the response is captured via the single one-dimensional combination $I(t) = I(v_1(t), v_2(t), v_3(t), ..., v_{2000}(t))$.

The following example is a modification of Math Function 2 in which there are ten underlying numerical variables $\{v_1(t), v_2(t), v_3(t), ..., v_{10}(t)\}$ associated with a single qualitative factor $t$, but their effects cannot be captured exactly by a two-dimensional $\mathbf{z}(t)$. The response function is

$$y(\mathbf{x}, v_1(t), v_2(t), ... v_{10}(t)) = \sum_{i=1}^{10} \frac{x_i v_{11-i}}{4000} + \prod_{i=1}^{10} \cos\left(\frac{x_i}{\sqrt{i}}\right) \sin\left(\frac{v_{11-i}}{\sqrt{i}}\right), \qquad (18)$$

where $-100 \leq x_i \leq 100$, and $-50 \leq v_i \leq 50, i = 1, ..., 10$. The qualitative factor $t$ has 5 levels, and we used the following mapping between $t$ and $\{v_1(t), v_2(t), ... v_{10}(t)\}$. We randomly generated all 50 values for $\{v_1(t), v_2(t), ... v_{10}(t): t = 1,2, ...,5\}$ uniformly within [-50, 50]. We used 20 replicates, and on each replicate, a different set of 50 values for $\{v_1(t), v_2(t), ... v_{10}(t): t = 1,2, ...,5\}$ were generated. For the experimental design, on each replicate we generated a different size-$n$ LHD in the $\{x_1, x_2, ... x_{10}\}$ space and then assigned the level for $t$ for each of the $n$ runs by randomly sampling one of its five levels. The BNGP model was fit to the same data as the LVGP model but using the underlying numerical $\{v_1(t), v_2(t), ... v_{10}(t)\}$ instead of $t$. Figure 10 shows the RRMSE comparison of the two models with different numbers of starting points for hyperparameter estimation (24 and 120) and training design sizes ($n = 80$ and 100). Even though the LVGP model uses a two-dimensional $\mathbf{z}(t)$, it achieved similar errors as the benchmark BNGP model and was evidently a reasonable approximation.

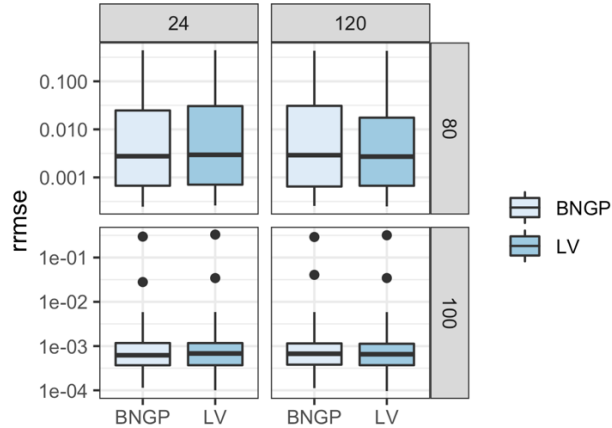

Figure 10: RRMSE comparison for the example in (18) across 20 replicates. The two columns represent 24 and 120 starting points for hyperparameter estimation, while the two rows represent design sizes $n = 80$ and 100. In all cases, our LVGP approach achieved similar errors as the BNGP approach that uses the underlying numerical $\{v_1(t), v_2(t), \ldots v_{10}(t)\}$.

*A Materials Design Example with Qualitative Inputs.* As emphasized throughout this paper, all qualitative factors in physics-based simulations must impact the response via some underlying numerical variables $\{v_1(t), v_2(t), v_3(t), \ldots\}$. However, in many situations, the underlying numerical variables may be so high-dimensional and the simulation physics so complex that it precludes conveniently identifying them and incorporating them into a GP model with only numerical variables. This is the case in the following materials design example (Balachandran et al. 2016). The data set consists of the simulated shear modulus (the response, $y$) of material compounds belonging to the family of M$_2$AX phases. The M atom has ten levels (i.e., ten different candidate choices for the compound) {Sc, Ti, V, Cr, Zr, Nb, Mo, Hf, Ta, W}, the A atom has two levels {C, N}, and the X atom has twelve levels {Al, Si, P, S, Ga, Ge, As,

Cd, In, Sn, Tl, Pb}. Thus, there are three qualitative factors with 10, 2, and 12 levels, respectively, to represent the different choices of atoms for the compound. Among the total 240 possible combinations, 17 combinations have negative shear modulus and thus are not considered in this example (see Balachandran (2016) for more details).

In the original study, the authors considered GP surrogate modeling. However, due to the high dimensionality and lack of transparency of the underlying $\{v_1(t), v_2(t), v_3(t), ...\}$, and due to the lack of effective GP modeling software for qualitative inputs, the authors used a GP model for numerical-only inputs with a relatively small set of numerical features (which can be viewed as a small subset of $\{v_1(t), v_2(t), v_3(t), ...\}$) that they suspected would have large impact on the response. In total, they chose seven features to serve as their numerical GP inputs, which are the *s*-, *p*-, and *d*-orbital radii for the M atom, and the *s*- and *p*-orbital radii for the A and X atoms. The orbital radii are from the Waber-Cromer scale. We refer to their GP modeling approach with only these seven features as numerical-only inputs as the "Quant_only" approach.

In the following, we show the advantages of using GP modeling with the original three qualitative inputs over the Quant-only GP model, and we also show the advantages of the LVGP model over existing GP models that can handle qualitative factors. We consider two versions of LVGP: One using only the three qualitative inputs (denoted LV_qual), and the other using the three qualitative inputs in addition to the seven orbital radii numerical variables (denoted LV). The seven numerical variables are in some sense redundant if the three qualitative inputs are included, since the latter are functions of the former. However, one might speculate that there may be advantages to including them along with the qualitative inputs if they truly have large impact on the response. The other three models that we compare are three existing GP models that we discussed in Section 3 to handle qualitative and quantitative inputs (ADD_UC, UC, and MC), all with the three qualitative inputs plus the seven numerical inputs.

There are 223 data points in total, and we used 200 of them for training and the remaining 23 to compute the test RRMSE. The training and test sets were chosen randomly from the 223 points, and we repeated this procedure for ten replicates, where on each replicate we chose different random subsets to serve as the training and test sets and repeated the modeling. Figure 11 shows that our LV method has much lower RRMSE than any of the other approaches (except LV_qual). Notice that Quant_only results in consistently large RRMSE, likely due to the seven chosen numerical features providing an insufficient quantitative representation of the effects of the qualitative levels. Although it includes the qualitative factors along with the quantitative features, the UC approach does not improve the accuracy compared to Quant_only. This is likely due to the fact that two of the qualitative variables have relatively large numbers of levels (10 and 12), resulting in a large number of parameters to estimate in the UC model. Both MC and ADD_UC have better RRMSE than UC, although our LV model achieves even better RRMSE. The best performing model was LV_qual, since its $25^{th}$, $50^{th}$, and $75^{th}$ RRMSE percentiles were all slightly better than those for the LV model, and substantially better than all other models. It is somewhat surprising that the LV_qual model performed better than the LV model, since the additional seven numerical features included in the LV model were speculated to have large effect. The benefit of including the additional seven numerical features appears to be offset by the additional challenge of estimating more hyperparameters. We view this as evidence that our LVGP approach handles qualitative factors efficiently enough that attempting to identify important numerical input features is not necessary.

This example illustrates an important reason why one would consider using qualitative factors in a GP model, even though their effects must be due to underlying numerical variables: Without definitive prior knowledge and a simulator whose mechanisms are transparent, selecting an appropriate set of low-dimensional features of the high-dimensional $\{v_1(t), v_2(t), v_3(t), ...\}$ is often subjective and provides an incomplete representation of the effects of the qualitative $t$. An LVGP model that includes the qualitative factors as inputs can account for the information not captured by quantitative variable features, thereby improving the GP model predictions.

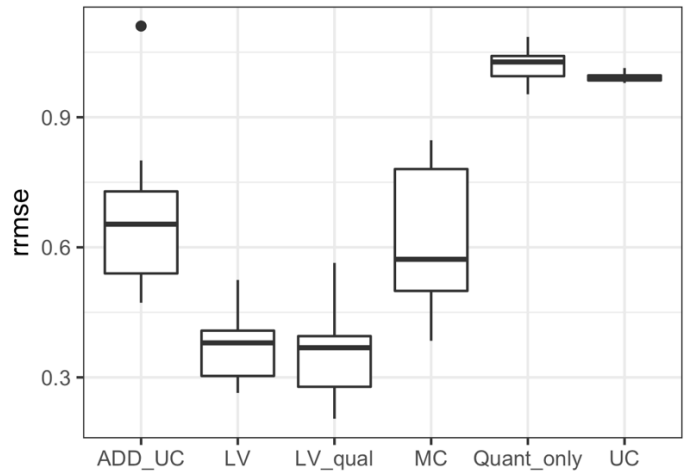

Figure 11: RRMSE comparison for the materials design example. Our LVGP method with (LV) or without (LV_qual) the additional seven numerical features achieved the lowest RRMSE.

## 6. CONCLUSIONS

In this article, we have developed a novel latent variable model for GP-based simulation response surface modeling with both quantitative and qualitative factors. The approach maps the qualitative factor levels to a corresponding set of 2D latent numerical variable values, so that distances in the latent variable space account for response correlations across levels of the qualitative factors. We have argued that the proposed two-dimensional latent variable model (6) is flexible enough to accurately capture complex correlations of many qualitative factors. To support this, we have (1) demonstrated consistently superior predictive performance across a variety of both mathematical and engineering examples (Figure 3, Figure 6, Figure 11) and (2) provided a physical explanation of why differences in the response behavior across qualitative factor levels are truly due to underlying numerical variables that can be mapped down to a lower dimensional space of latent variables (e.g., the beam bending example in Section 4).

Another desirable characteristic of our latent variable approach is that the estimated latent variables provide insight into the relationship between the levels of a factor, regarding how similar or different the response surfaces are for the different levels. In all of our examples, visualization of the latent variable space (Figure 4, Figure 7, and Figure 8) has successfully revealed the structure of the true underlying variables that account for the response differences between levels. Moreover, in contrast to the existing methods for handling qualitative factors that were reviewed in Section 3, our latent variable approach is compatible with any standard GP correlation function, including nonseparable correlation functions such as power exponential, Matèrn and lifted Brownian. This allows greater flexibility when modeling complex systems. The resulting covariance function in our latent variable model always results in a valid (nonnegative definite) covariance matrix without having to incorporate additional constraints, making the MLE routine easier to implement.

# ACKNOWLEDGEMENTS

This work was supported in part by National Science Foundation Grant NO. CMMI-1537641, DMREF Grant NO. 1729743, and the NIST-ChiMaD (Center for Hierarchical Materials Design) Grant.